# Bayesian Assessment of a Connectionist Model For Fault Detection

Stephen I. Gallant*

College of Computer Science
Northeastern University
Boston, Ma. 02115 USA

## Abstract

A previous paper [2] showed how to generate a linear discriminant network (LDN) that computes likely faults for a noisy fault detection problem by using a modification of the perceptron learning algorithm called the pocket algorithm. Here we compare the performance of this connectionist model with performance of the optimal Bayesian decision rule for the example that was previously described. We find that for this particular problem the connectionist model performs about 97% as well as the optimal Bayesian procedure.

We then define a more general class of *noisy single-pattern boolean (NSB) fault detection problems* where each fault corresponds to a single pattern of boolean instrument readings and instruments are independently noisy. This is equivalent to specifying that instrument readings are probabilistic but conditionally independent given any particular fault. We prove:

1. The optimal Bayesian decision rule for every NSB fault detection problem is representable by an LDN containing no intermediate nodes. (This slightly extends a result first published by Minsky & Selfridge.)

2. Given an NSB fault detection problem, then with arbitrarily high probability after sufficient iterations the pocket algorithm will generate an LDN that computes an optimal Bayesian decision rule for that problem.

In practice we find that a reasonable number of iterations of the pocket algorithm produces a network with good, but not optimal, performance.

**Keywords:** pocket algorithm, connectionist model, Bayesian model, fault detection, pattern recognition, classification

## 1 Introduction

A previous paper [2] derived a connectionist model for a fault detection problem, but failed to compare the performance of that model with optimal Bayesian performance. This omission stemmed from the paper's focus on how easy it was to construct an expert system for such a noisy and redundant problem. However since the example

*Partially supported by National Science Foundation grant IRI-8611596.

given in that paper can be analyzed probabilistically it is important to make such a comparison.

More generally it is interesting to contrast connectionist learning algorithms with probabilistic approaches since both paradigms are experiencing a renewal of popularity. The following analysis represents one sample point—and only one sample point—for such a comparison. Note that the connectionist model was developed and its details presented before any probabilistic analysis was performed. This makes it somewhat more suitable for a comparison experiment.

The next section briefly reviews the "Lemonade" fault detection task of [2] and develops a probabilistic analysis for comparison. We then prove relationships between Bayesian and connectionist models for a more general class of fault detection (or pattern recognition) problems. For general background on connectionist expert systems see [3].

## 2 The "Lemonade" Fault Detection Problem

Figure 1 below summarizes the important points of the lemonade fault detection problem from [2]. There are 9 possible faults $\{G_1, \cdots, G_9\}$, and the model assumes that exactly one of them is present at any time. Fault $G_9$ corresponds to lack of any fault, i.e. normal operation of the system.

| Failure Mode | Frequency | Importance | Final Ratio | \multicolumn{8}{c}{A Matrix — Noise-free Instrument Readings} |||||||||
|---|---|---|---|---|---|---|---|---|---|---|---|
| | | | | V1 | V2 | V3 | V4 | V5 | V6 | V7 | V8 |
| G1: | 1 | 20 | 20 | 1 | 1 | −1 | 1 | 1 | 1 | 1 | 1 |
| G2: | 1 | 2 | 2 | −1 | −1 | 1 | 1 | 1 | 1 | 1 | 1 |
| G3: | 1 | 2 | 2 | −1 | −1 | −1 | 1 | 1 | 1 | 1 | 1 |
| G4: | 1 | 2 | 2 | −1 | −1 | −1 | −1 | −1 | −1 | −1 | 1 |
| G5: | 1 | 2 | 2 | −1 | −1 | −1 | 1 | 1 | −1 | −1 | 1 |
| G6: | 1 | 2 | 2 | −1 | −1 | −1 | −1 | −1 | 1 | 1 | −1 |
| G7: | 2 | 2 | 4 | −1 | −1 | −1 | −1 | 1 | −1 | −1 | 1 |
| G8: | 2 | 2 | 4 | −1 | −1 | −1 | −1 | −1 | −1 | 1 | −1 |
| G9: | 40 | 1 | 40 | −1 | −1 | −1 | −1 | −1 | −1 | −1 | −1 |
| Total Examples: | | | 78 | 15% | 25% | 20% | 15% | 10% | 20% | 10% | 5% |
| | | | | \multicolumn{8}{c}{← Noise →} |||||||||

Figure 1: The Lemonade Fault Detection Problem

The *frequency* column gives relative frequencies for the faults; these correspond to prior probabilities. The *importance* column lists the penalty for failing to detect a particular fault when that fault is present. (This is not the most general utility function since it assumes the same penalty regardless of which incorrect fault was

128

chosen.) The *final ratio* column is the product of frequency and importance. We are able to take priors into account and to maximize our utility function by simply minimizing the probability of misclassification when examples are chosen according to this ratio.

Fault diagnosis is performed on the basis of 8 Boolean instruments, $\{V_1, \cdots, V_8\}$. The $A$ ("actual") matrix gives instrument readings for ny fault in the absence of noise. We assume the instruments are independently noisy, with noise values given along the bottom of figure 1. Thus if fault $G_1$ is present then instrument $V_1$ would read $+1$ with probability .85,
instrument $V_2$ would read $+1$ with probability .75,
instrument $V_3$ would read $-1$ with probability .8, etc.

Note that all of the information in the figure is potentially relevant, and that interactions can be complex. The connectionist learning approach described in [2] had the task of constructing an expert system for this problem without any direct access to the information given in figure 1; only training examples generated according to this model were available to the learning algorithm.

## 3 Probabilistic Analysis

A boolean fault detection problem uses a set of boolean instrument readings

$$V = \langle V_1, \cdots, V_n \rangle,$$

where each $V_j = \pm 1$ corresponding to *true* or *false*. For now we assume all readings are known; later we relax this assumption to allow unknown readings ($V_j = 0$). The possible failure modes are $\{G_1, \cdots, G_m\}$ with corresponding priors $\{P(G_i)\}$. We assume that the utility function has previously been folded into $P(G)$ as was done with the lemonade problem so that our goal is to maximize correct classifications based upon P(G) as a prior.[1]

By Bayes' Rule, the most likely failure mode $G^V$ given a set of instrument readings $V$ and taking importance into account is given by the $i \in [1 \ldots m]$ that maximizes

$$P(G_i|V) = \frac{1}{P(V)} \{P(V|G_i)P(G_i)\}. \tag{1}$$

Therefore an optimal Bayesian decision rule for selecting $G_i$ given $V$ is to select the fault $G_i$ that maximizes

$$P(V|G_i)P(G_i) \tag{2}$$

with ties in the maximum broken arbitrarily.[2]

This will give us an expected utility of

$$\mathbf{E}(P(G^V|V))$$

---

[1]This is possible only if the penalty for an incorrect diagnosis is independent of which fault was mistakenly selected.

[2]We could refer to *the* optimal Bayesian decision rule if there were no ties for maximum.

where

$$\begin{aligned}
\mathbf{E}(P(G^V|V)) &= \sum_V P(G^V|V)P(V) \\
&= \sum_V \frac{1}{P(V)} \max_i \{P(V|G_i)P(G_i)\} P(V) \quad \text{from (1)} \\
&= \sum_V \max_i \{P(V|G_i)P(G_i)\} \quad (3)
\end{aligned}$$

Note that the "figure of merit" used in [2] is this expected utility multiplied by 1000.
To compute $P(V|G_i)$ we first define noise-free patterns by:

$A_{i,j} = \pm 1$ according to the entry for $G_i$ and $V_j$ given in figure 1. (These are *actual* readings before noise is applied.)

and define noise by:
$$\mathcal{N}_{i,j} = P(V_j \neq A_{i,j}|G_i).$$

Then the probability that instrument reading $j$ will take on value $V_j$ when fault $G_i$ occurs is given by

$$P(V_j|G_i) = \begin{cases} 1 - \mathcal{N}_{i,j} & \text{if } A_{i,j} = V_j \\ \mathcal{N}_{i,j} & \text{otherwise.} \end{cases} \quad (4)$$

Finally we can compute
$$P(V|G_i) = \prod_j P(V_j|G_i) \quad (5)$$

because the $V_j$'s are conditionally independent given $G_i$ according to the noise generation model. Note that for the lemonade problem in [2] noise does not depend upon the underlying fault so that

$$\mathcal{N}_{1,j} = \mathcal{N}_{2,j} = \mathcal{N}_{m,j} = noise_j$$

where $noise_j$ is listed in figure 1. For example,

$$P(V = \langle 1, -1, 1, 1, 1, 1, 1, 1 \rangle | G_1) = \\ (.85)(.25)(.20)(.85)(.90)(.80)(.90)(.95).$$

## 4 Comparison of Network Performance to Bayesian Optimum

We computed the optimal expected utility from (3) for the lemonade problem, a calculation involving 256 maximizations of 9 computed quantities. The resulting optimal expected utility was .8278.

We then checked our calculations and gathered some additional statistics as follows. We produced 5000 additional noisy training examples (using a new random seed) and computed the most likely $G_i$ by the connectionist network and by the optimal Bayesian rule (2). The Bayesian rule was correct for 4119 cases giving an average utility of .8238. This was in close agreement with the theoretical .8278. In 4399 cases (87%) both methods gave the same answer. The comparison matrix was:

|  | Bayes correct | Bayes wrong |
|---|---|---|
| Network correct | 3830 | 200 |
| Network wrong | 289 | 681 |

For these runs the network model was correct for 4030 cases, giving an average utility of .806. (The average utility quoted in [2] was "approximately .810.") The optimal Bayes figure is clearly better than the network's but the difference is small. The ratio of performance is about $.806/.8278 = 97\%$. Thus the connectionist network does almost as well as the optimal Bayes approach.

Note however that the connectionist network was generated solely from training examples *without any information about the underlying process* while the optimal Bayesian decision rule was constructed using this knowledge. Thus a standard probabilistic approach based solely upon the examples that were used for network generation might easily do worse than the network (and almost certainly would do worse if its prior selection of underlying models was poorly matched to the problem at hand).

We were initially surprised that the connectionist network's performance was so close to optimal Bayes performance. However the following theorems show we should not have been so surprised.

## 5 Noisy Single-pattern Boolean Fault Detection Problems

Define a *noisy single-pattern boolean (NSB) fault detection problem* $\chi = \{G, V, P, A, \mathcal{N}\}$ as

- a set of faults $G = \{G_i, i = 1, \ldots, m\}$ with priors $P(G_i)$. (The priors may subsume importance information as with the lemonade problem.)

- instrument readings consisting of boolean n-vectors, $V = \langle V_j \rangle$, where $V_j = \pm 1, j = 1, \ldots, n$.

- a set of $m$ $n$-dimensional pattern vectors $A_{i,*}$ where $A_{i,j} = \pm 1$. $A_{i,*}$ is the single pattern that corresponds to fault $G_i$ in the absence of noise.

- noise probabilities $0 \leq \mathcal{N}_{i,j} \leq 1/2$ where $\mathcal{N}_{i,j}$ is the probability that $V_j$ differs from $A_{i,j}$ for fault $G_i$. Note that instrument noise is considered to be independent for any given fault.

Equivalently we could view NSB fault detection problems as multi-valued classification problems involving probabilistic boolean features, where the features are conditionally independent given a particular classification. Then we could define

$$A_{i,j} = \begin{cases} +1 & \text{if } P(V_j = +1|G_i) \geq 1/2 \\ -1 & \text{otherwise,} \end{cases}$$
$$\mathcal{N}_{i,j} = 1 - P(V_j = A_{i,j}|G_i).$$

131

Clearly NSB fault detection problems can also be viewed as a class of pattern recognition problems. For example we might consider the problem of deciding which of $m$ images is present upon a retina of binary points where noise is present. Similarly we may be interested in diagnosing which of $m$ diseases is present given a set of $n$ possible symptoms, where each disease is associated with a subset of symptoms and noise is present in the symptoms.[3]

Finally we define a class of connectionist models $C_n^m$ as the class of *linear discriminant network models with $n$ inputs, no intermediate cells, and $m$ output cells where the output cells form a* **choice** *group (also called a "winner-take-all" group).* In a $C_n^m$ network the only output cell to assume an activation of $+1$ is the cell with the largest[4] weighted sum of its inputs; all other output cells take on activations of $-1$. (Nilsson calls such networks *linear machines* [7].) See [2] for more detail.

The following Theorem says that any NSB fault detection problem has an optimal Bayesian decision rule that is representable by a $C_n^m$ network. Nilsson credits J. W. Jones for the essential idea, but Minsky & Selfridge [6,5] first published it. Nilsson [7] and Duda & Hart [1] also presented versions. The theorem we give incorporates the noise-free case where $\mathcal{N}_{i,j}$ is allowed to be 0. We employ the common notation $|\{\cdots\}|$ to denote the size of the set within brackets.

**Theorem 1** *Given an NSB fault detection problem $\chi = \{G, V, P, A, \mathcal{N}\}$ and any numbers $\alpha, \beta$ where $\alpha > 0$, then there exists a $C_n^m$ network, $C(\alpha, \beta)$, that computes the optimal Bayesian decision rule for $\chi$ where weights and biases for $C(\alpha, \beta)$ are given by:*

*For $j > 0$:*

$$w_{i,j} = \begin{cases} \alpha A_{i,j} \log\left(\frac{1-\mathcal{N}_{i,j}}{\mathcal{N}_{i,j}}\right), & \text{if } \mathcal{N}_{i,j} > 0 \\ \alpha A_{i,j} K, & \text{if } \mathcal{N}_{i,j} = 0 \end{cases} \quad (6)$$

*bias:*

$$w_{i,0} = \beta + \alpha \left\{ 2 \log \left( P(G_i) + \prod_{j\,:\,\mathcal{N}_{i,j} > 0} (1-\mathcal{N}_{i,j})(\mathcal{N}_{i,j}) \right) - K \left|\{j : \mathcal{N}_{i,j} = 0\}\right| \right\} \quad (7)$$

*and $K$ is any constant greater than*

$$\max_i \left\{ |\log P(G_i)| + \sum_{j\,:\,\mathcal{N}_{i,j} > 0} |\log((1-\mathcal{N}_{i,j})(\mathcal{N}_{i,j}))| + \sum_{j\,:\,\mathcal{N}_{i,j} > 0} \left| A_{i,j} \log\left(\frac{1-\mathcal{N}_{i,j}}{\mathcal{N}_{i,j}}\right) \right| \right\}$$

*proof:*

---

[3] Conditional independence of symptoms given a disease is not realistic in general. However if we consider associated symptom *groups* rather than lower level symptoms then conditional independence of groups given a disease seems more plausible.

[4] Ties are broken arbitrarily.

Omitted. □

Note that for the lemonade problem we can simplify the bias weights to:

$$w_{i,0} = \beta + 2\alpha \log P(G_i)$$

since $\mathcal{N}_{i,j} = \text{noise}_j > 0$.

The above weights agree with our intuitions about NSB problems in several respects. First $A_{i,j}$ agrees in sign with $w_{i,j}$, which is reasonable if we interpret the input as a noisy version of $A_{i,j}$. If $\mathcal{N}_{i,j} = 1/2$ then $w_{i,j} = 0$; in other words we ignore an input that adds no information due to noise. Similarly if $\mathcal{N}_{i,j}$ is small then $|w_{i,j}|$ is large; we pay heed to reliable inputs.

For the remainder of this paper we assume all probabilities and noise parameters are rational numbers.

## 6 Convergence Theorem

The pocket algorithm used for network generation is described in [4] and [3] and extended to $\mathcal{C}_n^m$ networks in [2]. The main theorem of this paper states that for NSB fault detection problems the pocket algorithm converges in probability to a $\mathcal{C}_n^m$ network that computes the optimal Bayesian decision rule.

**Theorem 2** *Given an NSB fault detection problem $\chi$ and given $\epsilon > 0$, there exists $N_0$ such that after $N > N_0$ iterations with probability $P > (1-\epsilon)$ the pocket algorithm will have produced weights for a $\mathcal{C}_n^m$ network that computes an optimal Bayesian fault classification for $\chi$.*

*proof:*

In theory we can generate a finite set of training examples $\mathcal{T} = \{\langle V, G_i \rangle\}$ that reflects $P(G)$, $\mathcal{N}_{i,j}$, and $P(V|G)$ since all probabilities are assumed rational. For the lemonade problem, $78(20)^8$ is a loose upper bound on $|\mathcal{T}|$. (In practice $\mathcal{T}$ is often too large to actually create.) Then generating training examples according to probabilities $P(G)$ and then adding noise to instrument readings is exactly equivalent to picking training examples at random from $\mathcal{T}$. Theorem 1 asserts the existence of a set of weights giving optimal Bayesian performance. By the Pocket Convergence Theorem [4], the pocket algorithm will produce a set of weights with at least equal performance to any set of weights with probability 1 as the number of iterations $N \to \infty$. Therefore these weights must give optimal Bayesian performance since it is impossible to do better. □

We have found that that large problems require too many iterations of the pocket algorithm to actually produce an optimal solution, but simulations indicate that the solutions that are generated are reasonably good (as was the case in the lemonade problem). We tried several classes of NSB problems to see how low we could make the relative performance. The worst performance of those we tried involved problems with 10 inputs and 20 possible faults where prior fault probabilities were constrained to be approximately equal. In this case the connectionist model did 85% as well as a theoretically optimal Bayesian model. (Note, however, that a Bayesian model generated from available training examples would not achieve the theoretical optimum performance.)

An interesting way of looking at the proof for Theorem 2 is to consider a black box generator of training examples for an arbitrary fault detection problem. If we use these training examples to generate a $C_n^m$ network, then (in the limit) that model would fit at least as well as the NSB fault detection model that *best* fits the given data.

The converse to Theorem 1 also holds:

**Theorem 3** *Given a $C_n^m$ network $C$, there exists an NSB fault detection problem $\chi$ satisfying*

1. $\mathcal{N}_{i,j} = 1/2$ only where $w_{i,j} = 0$
2. $A_{i,j}$ is $+1$ $\{-1\}$ if $w_{i,j}$ is positive $\{negative\}$
3. an optimal Bayesian decision rule for $\chi$ is given by the network $C$

The first two items prevent trivial solutions consisting of NSB fault detection problems where all faults are equally likely for every set of instrument readings.
*proof:*

The formulas for the weights given in Theorem 1 are invertible. We set $\alpha = 1$, then solve for $A_{i,j}$ and $0 < \mathcal{N}_{i,j} \leq 1/2$ for $j > 0$, and then solve for $\beta$ and $P(G_i)$ to satisfy (7) and $\sum P(G_i) = 1$. Details omitted. $\square$

We have now established a 1-1 correspondence between all NSB fault detection problems and all $C_n^m$ networks. Each fault detection problem is solved optimally by a network and for each network there exists an NSB problem for which it gives an optimal Bayesian decision rule. This correspondence can help our intuition in both domains.

## 7 Partial Information

Up to now we have required that all instrument readings $V_j$ be known. We would like to relax this assumption to be able to compute optimal $G_i$ when some of the readings are unknown. There are several ways to do this that we will only mention briefly here.

First, if all $\mathcal{N}_{i,j} > 0$ then Theorems 1-3 still hold if we set $V_j = 0$ for unknown instrument readings. In other words, the network model constructed in Theorem 1 will give correct Bayesian choices from partial information if no variable is noise-free.

Alternatively we can define our fault detection model so that we only permit responses corresponding to $G_i$ for which all noise-free readings are known to be

134

satisfied.[5] This assumption also extends Theorems 1-3 to cover the case of partial information.

Finally we could add an extra input to each cell for each instrument reading that takes on the value +1 if that reading is known and −1 otherwise. Appropriate weights can now be assigned to extend Theorems 1-3 to cover the case where partial information is present.

## 8 Conclusion

We have calculated for the particular lemonade problem discussed in [2] that the connectionist expert system approach produced a solution very close to optimal. More generally we have shown that the family of optimal decision rules for noisy single-pattern boolean fault detection problems are in correspondence with linear discriminant networks having no intermediate nodes. Moreover we have proven that the pocket algorithm produces sets of weights for an LDN that converges in probability to weights that give optimal Bayesian decision rules for the class of NSB fault detection problems.

It is interesting that NSB fault detection problems present a large class of non-separable problems for which we can compute optimal solutions for the corresponding linear discriminant problems (for $n \leq 20$). Therefore this class might serve as convenient test data to evaluate the actual performance of the pocket algorithm or other algorithms.

Finally it would be interesting to obtain analytical or empirical results on convergence speed of the pocket algorithm for this class of fault detection problems.

**Acknowledgment:** Thanks to Peter Cheeseman for motivating this study and to Max Henrion and the referees for useful suggestions.

---

[5] If no such $G_i$ exists, than the choice of $G_i$ is arbitrary.